\documentclass[sn-mathphys-num]{sn-jnl}

\usepackage{graphicx}%
\usepackage{multirow}%
\usepackage{amsmath,amssymb,amsfonts}%
\usepackage{amsthm}%
\usepackage{natbib}%
\usepackage{upgreek}%
\usepackage{textgreek}%
\usepackage{mathrsfs}%
\usepackage[title]{appendix}%
\usepackage{xcolor}%
\usepackage{textcomp}%
\usepackage{manyfoot}%
\usepackage{booktabs}%
\usepackage{algorithm}%
\usepackage{algorithmicx}%
\usepackage{algpseudocode}%
\usepackage{listings}%
\usepackage{gensymb}%
\usepackage[colorinlistoftodos]{todonotes}%
\usepackage{soul}%
\usepackage{float}
\usepackage{subcaption}

\begin{document}

\title[Article Title]{Differentiable Jitter Correction using Deep Learning-based Image Quality Metric for Phase-Contrast Micro-CT}


\author*[1,2]{\fnm{Junan} \sur{Chen}}\email{chen@imfusion.com}

\author[2]{\fnm{Yiting} \sur{Jia}}

\author[1]{\fnm{Joscha} \sur{Maier}}

\author[2]{\fnm{Dominik} \sur{John}}

\author[2,3]{\fnm{Sami} \sur{Wirtensohn}}

\author[4]{\fnm{Imke} \sur{Greving}} 

\author[4]{\fnm{Silja} \sur{Flenner}}

\author[1]{\fnm{Matthias} \sur{Wieczorek}} 

\author[2]{\fnm{Julia} \sur{Herzen}}

\affil[1]{\orgname{ImFusion GmbH}, \orgaddress{\street{Ridlerstr. 57},  \postcode{80339} \city{Munich}, \country{Germany}}}

\affil[2]{\orgdiv{Research Group Biomedical Imaging Physics, Department of Physics}, \orgname{TUM School of Natural Sciences and Munich Institute of Biomedical Engineering, Technical University of Munich}, \orgaddress{\street{James-Franck-Str. 1}, \postcode{85748} \city{Garching}, \country{Germany}}}

\affil[3]{\orgdiv{Centre for X-Ray and Nano Science CXNS}, \orgname{Deutsches Elektronen-Synchrotron (DESY)}, \orgaddress{\street{Notkestr. 85}, \postcode{22067} \city{Hamburg}, \country{Germany}}}

\affil[4]{\orgdiv{Institute of Materials Physics}, \orgname{Helmholtz-Zentrum Hereon}, \orgaddress{\street{Max-Planck-Str. 1}, \postcode{21502} \city{Geesthacht}, \country{Germany}}}

\abstract{X-ray phase-contrast micro computed tomography is an advanced imaging modality capable of revealing the structural and material properties of soft tissue at micrometer-scale resolution. However, the high spatial resolution that makes this technique advantageous also imposes stringent requirements on the calibration of the imaging setup, its mechanical stability, and the sample's stability. Even minor geometric errors have a substantially greater effect on reconstruction quality than in conventional CT. Performing reconstruction without accounting for these instabilities introduces artifacts, such as blurring and streaking, which substantially degrade image quality. This paper proposes a fully differentiable iterative correction method using a deep learning-based image quality metric that estimates and compensates per-projection rigid jitter directly from the acquired projection data, without a pre-scan motion-free reference. The approach builds on a gradient-based auto-focus strategy adapted to parallel-beam geometry. A set of candidate objective functions is benchmarked in a controlled study, and the sensitivity of the visual information fidelity (VIF) metric to the jitter artifact is verified with the target phase-contrast data. To operate without a clean reference, a compact 3D convolutional neural network is trained to predict the VIF score from a single corrupted volume. A spatially selective total variation penalty applied exclusively to the image background is introduced to penalize spurious high-frequency structures that otherwise emerge during optimization. Experiments on biological specimens acquired at different synchrotron beamlines are conducted.  Evaluation uses jitter motion applied to simulated and experimentally acquired projection data. The result confirms that the integrated pipeline reliably recovers fine structural detail lost due to jitter, with generalization demonstrated across morphologically distinct samples.}

\keywords{Deep Learning, Motion Compensation, Motion Metric, Phase-contrast CT, Micro-CT}

\maketitle

\section{Introduction}\label{sec1}

Among the modalities available for high-resolution 3D imaging, X-ray phase-contrast micro computed tomography (\textmu CT) stands out as an exceptionally sensitive tool for visualizing soft tissue at micrometer-scale resolution. In contrast to conventional CT, which exploits differences in X-ray absorption, phase-contrast \textmu CT leverages refraction and interference phenomena to achieve markedly superior contrast for weakly absorbing biological structures, enabling medical and biological investigations at scales previously inaccessible to standard clinical systems \cite{desyphasecontrast}. This increased image quality comes, however, at the cost of considerably tighter operational constraints. Because phase-contrast \textmu CT reveals structural information at spatial frequencies much higher than conventional CT, even geometric inconsistencies at the micrometer scale, which are negligible in clinical scanning, can propagate into the reconstructed volume as significant artifacts. Two sources of such inconsistency are particularly relevant in practice: sample drift or jitter, driven by the inherently long acquisition time associated with high-resolution scanning, and device jitter, whose impact grows as voxel dimensions shrink to the micrometer range. The instabilities in a phase-contrast \textmu CT setup are stochastic in nature and persist throughout the scan, meaning that even with synchronization-based hardware measures in place, residual misalignments can still degrade image quality \cite{DeMarco:18, Brombal2021-ho, Odstrcil:19}. When left unaddressed, these inter-projection misalignments manifest in the reconstruction as blurring of fine microstructural features or streak-like intensity variations that obscure the very structural information that the modality is designed to reveal \cite{Boas2012}, hence affecting the outcome of downstream tasks such as cross-correlation with or mapping to a second modality \cite{john2026quantitative, chen2025deformable}.

Methods for CT geometry calibration and motion compensation are widely studied in clinical applications, spanning a broad spectrum of approaches: from methods leveraging external motion tracking signals (e.g., cardiac CT gated with electrocardiography (ECG) \cite{Desjardins2004}), to prior-knowledge-based approach that eliminates the need for gating by directly predicting displacement vector fields (DVFs) using a U-net-based model \cite{Maier2025}, to image-quality-driven methods such as auto-focus motion compensation, which optimizes projection geometry with respect to a predefined motion metric \cite{Sisniega2017}. Auto-focus motion compensation treats correction as an iterative optimization over the acquisition geometry. An image quality criterion guides the search for parameters that best restore reconstruction quality. Sisniega et al. applied auto-focus correction to extremity cone-beam CT using a sharpness metric and the derivative-free CMA-ES optimizer, which iteratively reconstructs the volume during the optimization for the motion model \cite{Sisniega2017}. Thies et al. embedded the full reconstruction pipeline into a differentiable graph, enabling gradient-based optimization and achieving convergence nearly an order of magnitude faster than CMA-ES \cite{Thies2025}. Phase-contrast \textmu CT routinely involves morphologically diverse specimens subject to unknown, per-projection jitter motion. Auto-focus methods require only the acquired data and an image-quality criterion, and the motion model can be specified independently of the metric. We therefore consider auto-focus methods and their variants to offer the greatest potential in this setting. However, hand-crafted auto-focus metrics such as sharpness and total variation (TV) can easily reach their limitation when attempting to correlate with the image degradation caused by various levels of motion. Huang et al. replaced the hand-crafted quality criterion in the auto-focus motion compensation with a learned surrogate trained to predict a visual information fidelity (VIF) \cite{Sheikh2006} score without a clean reference; their framework was demonstrated on human head CT \cite{Huang2022}. However, the motion metrics proposed in this context were originally developed for conventional CT. To our knowledge, a learned, reference-free image-quality criterion has not previously been used to drive jitter correction in X-ray phase-contrast \textmu CT. Furthermore, existing metrics are typically tailored to specific anatomical structures or regions, such as the human abdomen or head, limiting their generalizability to arbitrary biological samples in phase-contrast \textmu CT.

In this work, we propose a method that uses differentiable gradient-based optimization with a learned, reference-free quality metric for jitter correction in parallel-beam phase-contrast \textmu CT, in which the projection geometry and specimen diversity differ substantially from those in clinical CT. We derive the gradient matrix of the reconstruction operator for parallel-beam geometry. We train a compact 3D convolutional neural network (CNN) tailored to jitter-induced motion artifacts to predict VIF from a single degraded volume, removing the need for a paired clean reference at inference time. To resolve hallucination failure modes in which certain motion trajectories generate spurious high-frequency patterns that artificially inflate the objective function, we append a TV penalty restricted to the image background for suppressing non-physical textures without smoothing genuine internal sample structure.

\section{Methods}\label{sec2}
\subsection{Differentiable CT geometry optimization}
In image-quality-driven CT geometry optimization, the cost function is usually illustrated as follows:
\begin{equation}
\mathrm{C}(\mathbf{P}) = \mathrm{Q}(\mu(\mathbf{P})) + \beta \mathrm{R}(\mathbf{P}),
\label{eq:cost_function}
\end{equation}
where $\mathrm{Q}$ is an image quality term and $\mu(\mathbf{P})$ denotes the reconstruction image using the current projection matrices $\mathbf{P}$. $\mathrm{R}$ is a regularization term with a factor $\beta$. The regularization term can either be encouraging certain trajectory conditions or another image-domain term to counteract a specific degeneracy of the primary quality criterion \cite{Sisniega2017, Huang2022, Capostagno2021}. In this work, the latter option is chosen. The cost function profile is often complex, with numerous local optima. Conventional methods have extensively used gradient-free optimization strategies, which do not impose stringent differentiability requirements on the components within the optimization framework, while allowing navigation of cost functions and maintaining robustness against premature convergence to local optima \cite{Sisniega2017}. However, gradient-free optimization strategies require more iterations and, consequently, more intermediate reconstruction steps, which makes convergence computationally more expensive. In contrast, gradient-based strategies offer a more direct path to the optimum but necessitate an explicit expression of the gradient. The computation of the gradient of this optimization problem has been complicated by the requirement to differentiate through the non-trivial and computationally demanding CT backprojection operator. 

To compute the gradient of the cost function stated in Eq. \ref{eq:cost_function}, a gradient chain is formulated as follows:
\begin{equation}
\nabla_{\mathbf{P}} \mathrm{C}(\mathbf{P})
=
\left(\frac{\partial \mu(\mathbf{P})}{\partial \mathbf{P}}\right)^{\mathsf{T}}
\nabla_{\mu}\mathrm{Q}(\mu(\mathbf{P}))
+
\beta \nabla_{\mathbf{P}} \mathrm{R}(\mathbf{P}).
\label{eq:gradient_chain}
\end{equation}
Here, $\nabla_{\mu}\mathrm{Q}(\mu(\mathbf{P}))$ is the per-voxel sensitivity of the image quality term to the reconstructed volume, i.e., \ $\partial \mathrm{Q}/\partial \mu[\mathbf{X}]$ for every voxel $\mathbf{X}$. The remaining difficulty lies in the Jacobian $\partial \mu(\mathbf{P}) / \partial \mathbf{P}$. Although $\mu(\mathbf{P})$ admits a closed-form expression, differentiable reconstruction operators have conventionally been constructed to carry gradients back into the measured projections alone, leaving the geometry parameters outside the computational graph~\cite{Thies2023}. Obtaining those derivatives numerically via finite differences instead means re-evaluating the objective once for every free parameter, while implementations that delegate the task to automatic differentiation scale poorly on GPU hardware and become impractical at the parameter counts arising in per-projection correction~\cite{Thies2023}. Instead, following Thies et al.\cite{Thies2025}, the reconstruction operator is modeled explicitly as a per-voxel, bilinearly interpolated sampling of the projection data, which admits an analytic backward formulation.

Let $\mathbf{P} = \{\mathbf{P}_a\}_{a=1}^{N} \subset \mathbb{R}^{3\times4}$ denote the set of $N$ projection matrices, where $a$ indexes the individual acquired projections, each associated with its own projection image $S_a$. For a parallel-beam geometry, a voxel with homogeneous world coordinate $\tilde{\mathbf{X}} = (X_0, X_1, X_2, 1)^{\mathsf{T}}$ is mapped onto detector coordinates $(u_a, v_a)$ at projection angle $a$ by the first two rows of $\mathbf{P}_a$, without perspective division:
\begin{equation}
u_a(\mathbf{X}) = \mathbf{P}_a^{(0,:)}\,\tilde{\mathbf{X}}, \qquad v_a(\mathbf{X}) = \mathbf{P}_a^{(1,:)}\,\tilde{\mathbf{X}}.
\label{eq:parallel_projection}
\end{equation}
Let $S_a$ denote the projection recorded at angle $a$ after ramp filtering along the detector row direction. The reconstruction at voxel $\mathbf{X}$ is then obtained by accumulating the bilinearly interpolated values of $S_a$ across all angles:
\begin{equation}
\mu(\mathbf{P})[\mathbf{X}] = \sum_{a=1}^{N} S_a\big(u_a(\mathbf{X}), v_a(\mathbf{X})\big).
\label{eq:backprojection_model}
\end{equation}
Differentiating Eq.~\ref{eq:backprojection_model} with respect to the entries of $\mathbf{P}_a$ and applying the chain rule through Eq.~\ref{eq:parallel_projection} yields, for the $j$-th column ($j=0,\dots,3$) of the two rows that affect the projection,
\begin{equation}
\frac{\partial \mu(\mathbf{P})[\mathbf{X}]}{\partial \mathbf{P}_a[0,j]} = \frac{\partial S_a}{\partial u}\bigg|_{(u_a(\mathbf{X}),\,v_a(\mathbf{X}))} \tilde{X}_j,
\qquad
\frac{\partial \mu(\mathbf{P})[\mathbf{X}]}{\partial \mathbf{P}_a[1,j]} = \frac{\partial S_a}{\partial v}\bigg|_{(u_a(\mathbf{X}),\,v_a(\mathbf{X}))} \tilde{X}_j.
\label{eq:voxel_jacobian}
\end{equation}
The partial derivatives $\partial S_a / \partial u$ and $\partial S_a / \partial v$ are computed once per angle as the numerical gradient of the filtered projection image along its two axes, and are subsequently bilinearly interpolated at the sub-pixel location $(u_a(\mathbf{X}), v_a(\mathbf{X}))$. 
Substituting Eq.~\ref{eq:voxel_jacobian} into Eq.~\ref{eq:gradient_chain} and accumulating the contribution of every voxel, the gradient of the cost function with respect to a single projection matrix $\mathbf{P}_a$ reduces to a $3\times4$ geometry gradient matrix:
\begin{equation}
\mathbf{G}_a
=
\nabla_{\mathbf{P}_a}\mathrm{C}(\mathbf{P})
=
\sum_{\mathbf{X}}
w(\mathbf{X})
\begin{bmatrix}
\dfrac{\partial S_a}{\partial u}\Big|_{(u_a(\mathbf{X}),v_a(\mathbf{X}))} \\[6pt]
\dfrac{\partial S_a}{\partial v}\Big|_{(u_a(\mathbf{X}),v_a(\mathbf{X}))} \\[6pt]
0
\end{bmatrix}
\tilde{\mathbf{X}}^{\mathsf{T}}
\;\in\; \mathbb{R}^{3\times4},
\label{eq:geometry_gradient_matrix}
\end{equation}
where $w(\mathbf{X}) = \nabla_{\mu}\mathrm{Q}(\mu(\mathbf{P}))[\mathbf{X}] + \beta\, \nabla_{\mathbf{P}}\mathrm{R}(\mathbf{P})[\mathbf{X}]$ combines the image quality and regularization contributions at voxel $\mathbf{X}$. 
The complete gradient $\nabla_{\mathbf{P}}\mathrm{C}(\mathbf{P}) = \{\mathbf{G}_a\}_{a=1}^{N}$ is then passed directly to a gradient-based optimizer to update the projection geometry at every iteration.

The regularization term $\mathrm{R}$ in Eq.~\ref{eq:cost_function} constrains the optimizer to physically plausible reconstructions. We instantiate $\mathrm{R}$ as a TV penalty. Because the raw TV of a volume scales with its spatial dimensions and can exceed the bounded range of the image-quality term by several orders of magnitude, it is normalized by the TV of the initial reconstruction, so that at iteration $i$:
\begin{equation}
\mathrm{R}_i = \frac{\mathrm{TV}_i}{\mathrm{TV}_0},
\label{eq:tv_norm}
\end{equation}
where $\mathrm{TV}_i$ is the total variation of the volume reconstructed with the current geometry estimate and $\mathrm{TV}_0$ that of the initial, uncorrected reconstruction. The resulting term is dimensionless and is invariant to volume size, so that a single weight $\beta$ transfers across datasets of differing resolution. Evaluated over the entire volume, however, this penalty does not distinguish spurious texture from genuine internal structure and therefore smooths the very microstructural detail the correction is intended to recover. We consequently consider a spatially selective variant, in which the penalty is restricted to a binary mask $\Omega_{\mathrm{bg}}$ of voxels lying outside the sample region:
\begin{equation}
\mathrm{R}_i^{\mathrm{bg}} = \frac{\mathrm{TV}_{i,\,\mathrm{bg}}}{\mathrm{TV}_{0,\,\mathrm{bg}}},
\qquad
\mathrm{TV}_{i,\,\mathrm{bg}} = \sum_{\mathbf{X} \in \Omega_{\mathrm{bg}}}
\big\|\nabla \mu(\mathbf{P})[\mathbf{X}]\big\|_1 ,
\label{eq:tv_norm_bg}
\end{equation}
so that non-physical textures in nominally empty space are suppressed while structure inside the specimen is left untouched. The sub-gradient of $\mathrm{R}$ with respect to the volume enters the per-voxel weight $w(\mathbf{X})$ of Eq.~\ref{eq:geometry_gradient_matrix} and is backpropagated to the projection matrices through the same Jacobian as the image-quality term.

\subsection{Deep learning-based reconstruction quality metric}
The image-quality term $\mathrm{Q}(\cdot)$ appearing in Eq.~\eqref{eq:cost_function} is crucial to the geometry optimization. The reliability and behavior of this term directly determine whether the optimization converges to a physically meaningful optimum. To this end, we first conduct a quantitative evaluation of various metrics, sharpness, TV, structural similarity index (SSIM), and VIF, to determine whether they correlate with the jitter-induced geometric error in the phase-contrast CT dataset. Similar to SSIM, VIF is a reference-based metric, requiring access to a motion-free ground-truth volume that is unavailable at deployment time. We address this by training a CNN, namely DL-VIF, that predicts the VIF score directly from such a single volume, without requiring a reference. VIF is preferred over SSIM as the regression target because SSIM's response is computed from local structural comparisons and is therefore strongly tied to the specific structural content of the sample, whereas VIF is derived from a natural-scene-statistics model of the human visual system and quantifies the amount of statistical image information lost through a distortion channel, making it a more structure-agnostic and information-theoretically grounded target~\cite{Sheikh2006}. The following subsections illustrate the method for metrics evaluation, network architecture, and training procedure.

\subsubsection{Metrics evaluation}\label{secsimulation}
To verify the robustness of the VIF quality metric for the auto-focus CT geometry optimization, we conduct a quantitative evaluation with a phase-contrast \textmu CT dataset of a C. Elegans worm acquired at the Hereon at DESY P05 nano-CT beamline \cite{Wirtensohn2026}. Different levels of random rigid transformation are introduced into the ground-truth projection matrices to simulate unknown jitter during the projection data acquisition. Motion-corrupted volumes are then obtained by reconstructing the projection data with contaminated projection matrices. 
\begin{equation}
\mu_{\mathrm{corrupted}} = \mu\!\left(\left\{\bar{\mathbf{P}}_a\, \mathbf{T}_a\right\}_{a=1}^{N}\right),
\label{eq:motion_corrupted_volume}
\end{equation}
$\bar{\mathbf{P}}_a \in \mathbb{R}^{3\times4}$ denotes the ideal, motion-free projection matrix of acquisition angle $a$, and $\mathbf{T}_a$ is a $4\times4$ homogeneous rigid-transformation matrix applied to that angle. Each $\mathbf{T}_a$ is sampled independently for each of the $N$ acquisition angles by a random rotation, parameterized by Euler angles $\boldsymbol{\phi}_a \sim \mathcal{U}(-\theta_{\max}, \theta_{\max})^3$, and a random translation $\mathbf{t}_a \sim \mathcal{U}(-s_{\max}, s_{\max})^3$; the severity parameters $\theta_{\max}$ and $s_{\max}$ jointly control the magnitude of the simulated rotational and translational jitter, respectively. Right-multiplying $\bar{\mathbf{P}}_a$ by $\mathbf{T}_a$ perturbs the geometry independently at every angle, so that the resulting set $\{\bar{\mathbf{P}}_a\,\mathbf{T}_a\}_{a=1}^{N}$ no longer describes the correct acquisition trajectory but one contaminated by unknown, per-angle motion. These contaminated matrices are passed into the same reconstruction operator $\mu(\cdot)$ used in Eq.~\ref{eq:cost_function}, and the resulting volume $\mu_{\mathrm{corrupted}}$ exhibits the blurring artifacts characteristic of jitter during acquisition. Various levels of translation and rotation are applied to generate motion-corrupted volumes. Four metric candidates are evaluated with reconstruction volumes with increasing motion level.
\subsubsection{Model architecture}
The DL-VIF network adapts the reference-free image-quality learning framework of Huang et al.~\cite{Huang2022} to a more lightweight, 3D architecture that emphasizes global image-quality assessment over the localization of local artifacts. Feature extraction relies on 3D residual blocks to enable stable training of a deeper network. Each block consists of a $1\times1\times1$ channel-mapping convolution followed by instance normalization, a $3\times3\times3$ spatial convolution (padding 1), a second instance normalization, and a parallel $1\times1\times1$ skip-connection that matches input and output dimensionality. The two paths are summed and passed through a Leaky ReLU activation, so that the block learns a residual correction to its input rather than a full mapping.

\begin{figure}[H]
    \centering
    \includegraphics[width=\linewidth]{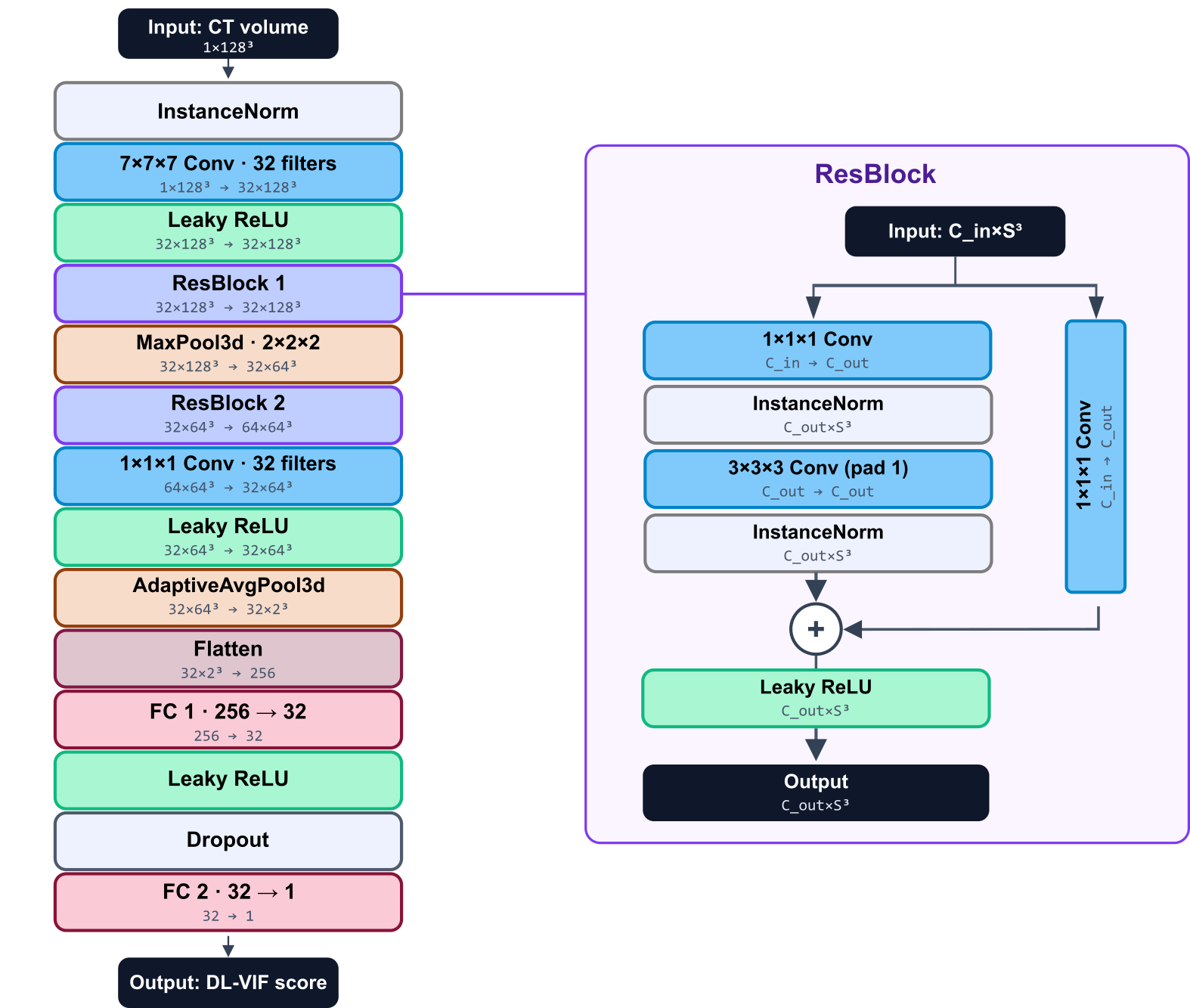}
    \caption{The network architecture of DL-VIF.}
    \label{fig:architecture}
\end{figure}

Figure~\ref{fig:architecture} shows the resulting pipeline, which maps a single-channel $128\times128\times128$ input volume to a scalar quality score. Since the network must be repeatedly evaluated and backpropagated at every iteration of the downstream gradient-based motion-correction optimization, a substantially larger input would make the per-iteration computation and activation-memory cost of both passes impractical. We therefore fix the input resolution at $128^3$. Coarse features are first extracted via instance normalization, followed by a $7\times7\times7$ convolution (32 filters), capturing the large-scale geometry of the volume. Two residual blocks then encode increasingly detailed features, separated by a $2\times2\times2$ max-pooling layer that halves the spatial resolution ($128^3 \rightarrow 64^3$) to control computational cost. A $1\times1\times1$ convolution after the second residual block consolidates channel-wise information, after which an adaptive average-pooling layer compresses the spatial extent from $64^3$ down to a $2\times2\times2$ grid, while the channel depth is increased to 32 to retain representational capacity. The resulting $32\times2\times2\times2$ tensor is flattened into a 256-dimensional vector and passed through two fully connected layers ($256 \rightarrow 32 \rightarrow 1$), yielding the predicted DL-VIF score.

This global-pooling design is a deliberate departure from Huang et al.~\cite{Huang2022}. Since motion in our setting corrupts the volume as a whole through global stochastic jitter, the network only needs to assess an overall degradation level rather than localize where artifacts occur. Retaining fine spatial detail before the fully connected layers under this objective would instead let the network memorize specimen-specific structural patterns from the limited number of training specimens, at the cost of generalizing to unseen samples. Compressing the spatial dimension to $2\times2\times2$ (while compensating with a wider channel depth of 32) removes most positional information and shrinks the weights of the first fully connected layer to a few thousand, forcing the network to prioritize global, structure-agnostic degradation statistics over local structural recognition.

\subsubsection{Datasets and training}
\begin{figure}[H]
\centering
\begin{subfigure}[b]{0.23\textwidth}
    \centering
    \includegraphics[width=\textwidth]{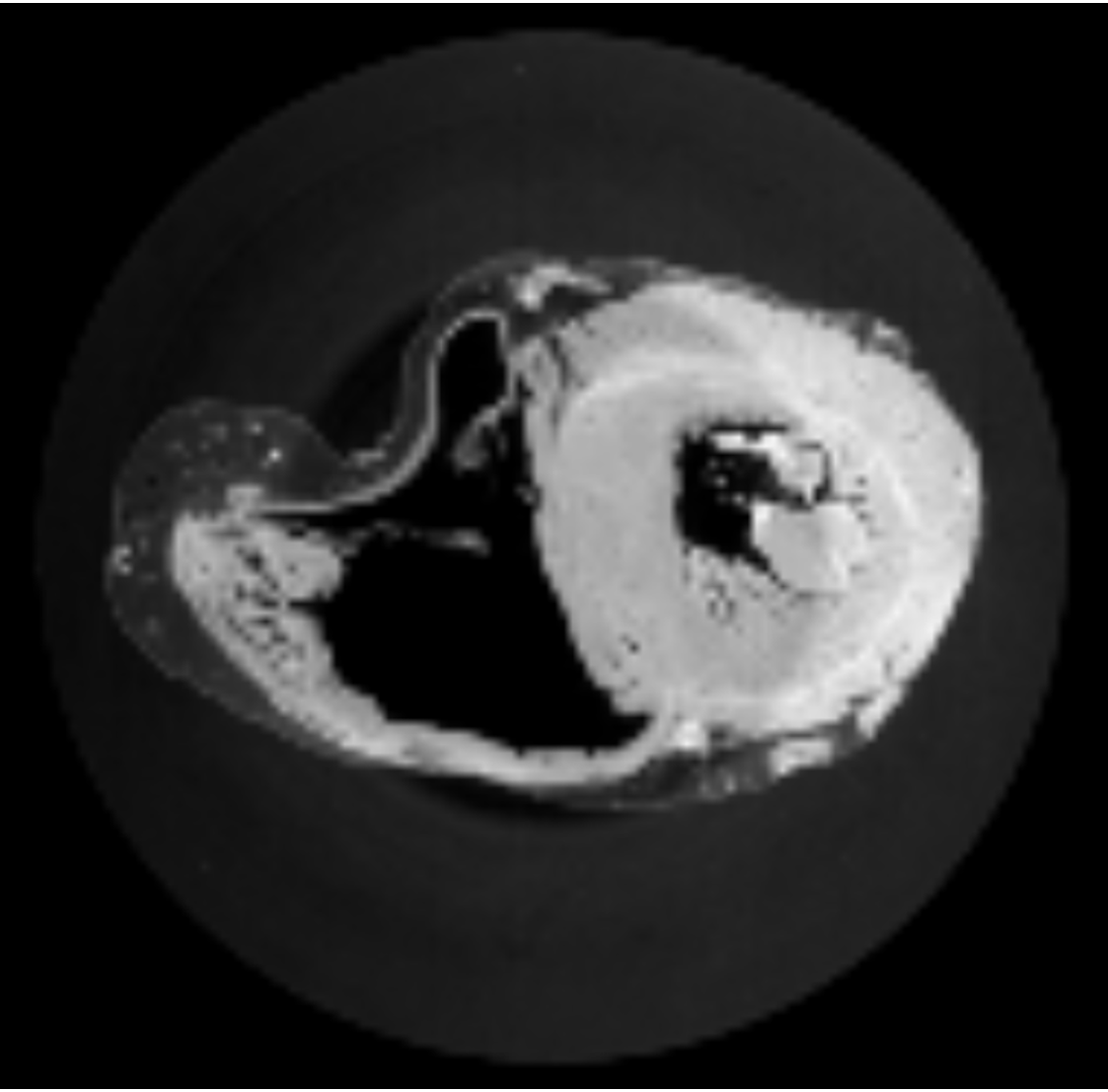}
    \caption{Heart 01}
\end{subfigure}
\hfill
\begin{subfigure}[b]{0.23\textwidth}
    \centering
    \includegraphics[width=\textwidth]{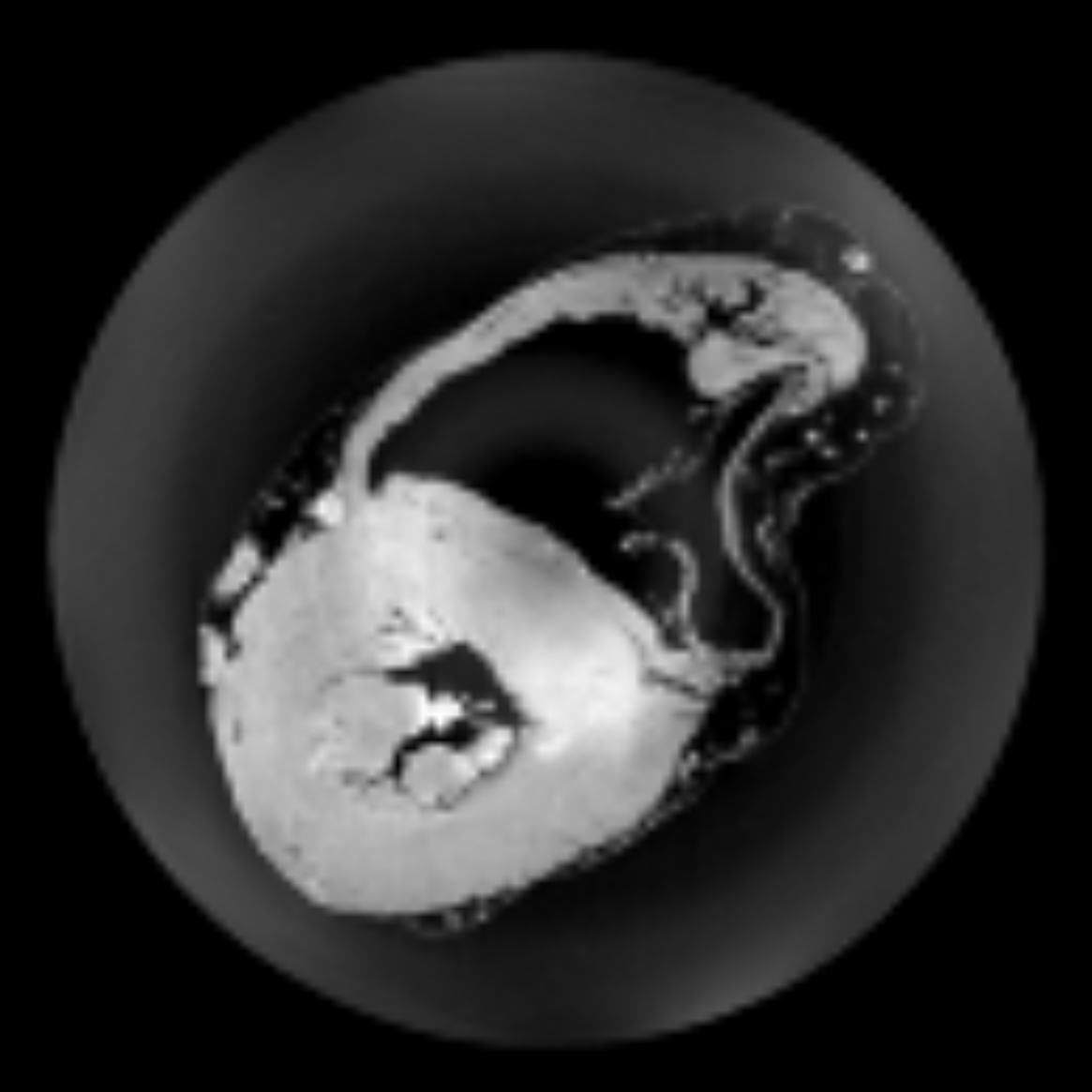}
    \caption{Heart 02}
\end{subfigure}
\hfill
\begin{subfigure}[b]{0.23\textwidth}
    \centering
    \includegraphics[width=\textwidth]{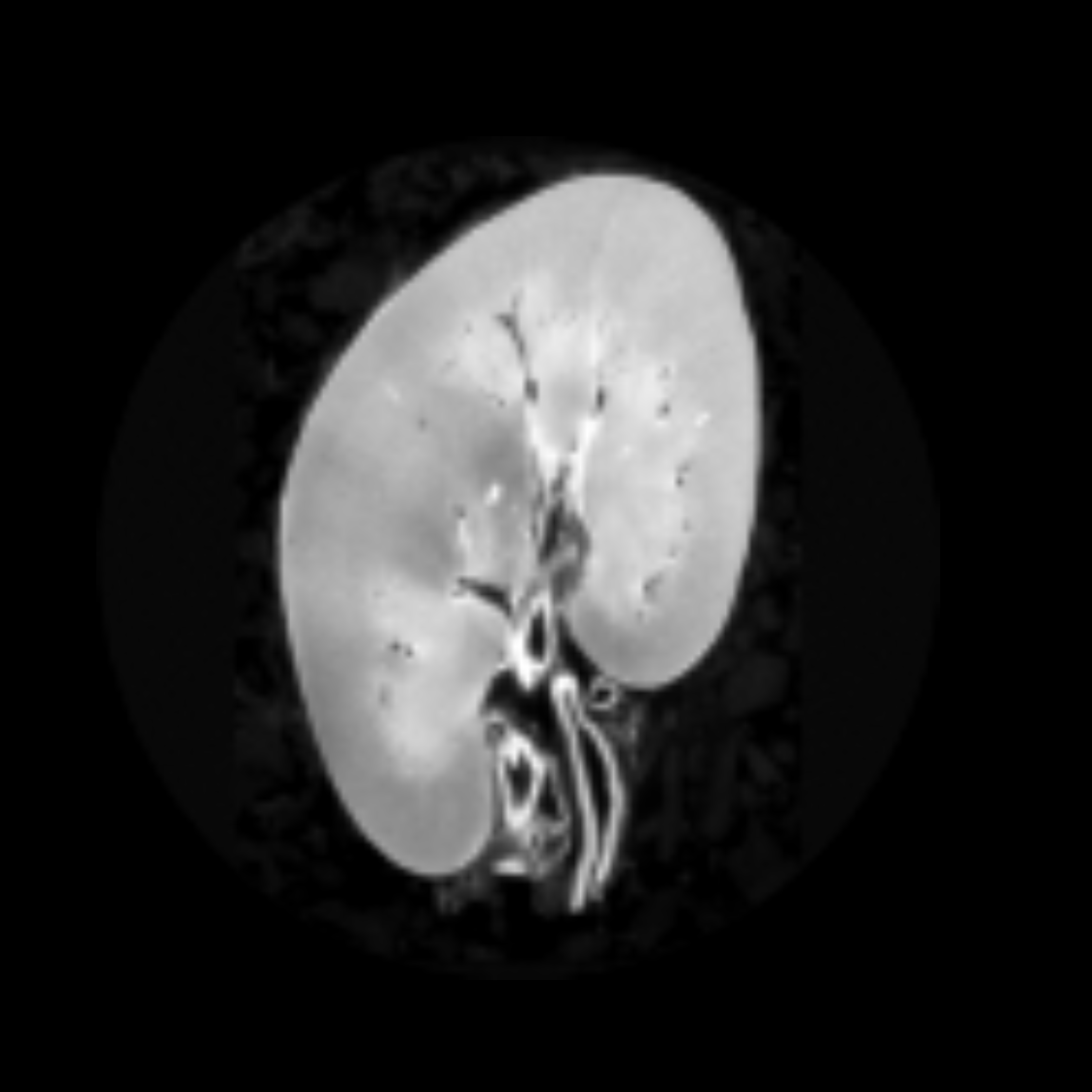}
    \caption{Kidney 01}
\end{subfigure}
\hfill
\begin{subfigure}[b]{0.23\textwidth}
    \centering
    \includegraphics[width=\textwidth]{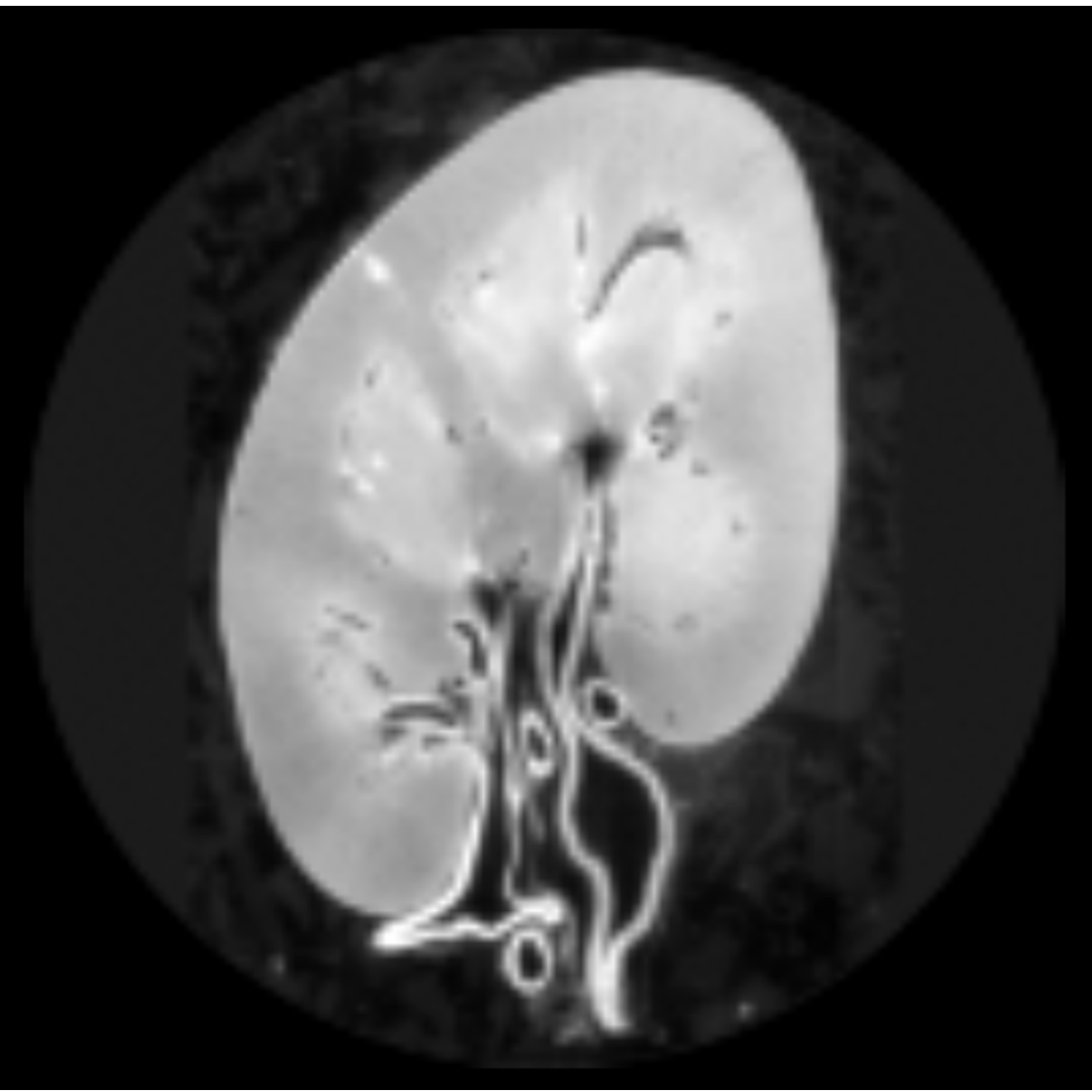}
    \caption{Kidney 02}
\end{subfigure}
\vspace{0.3cm}
\begin{subfigure}[b]{0.23\textwidth}
    \centering
    \includegraphics[width=\textwidth]{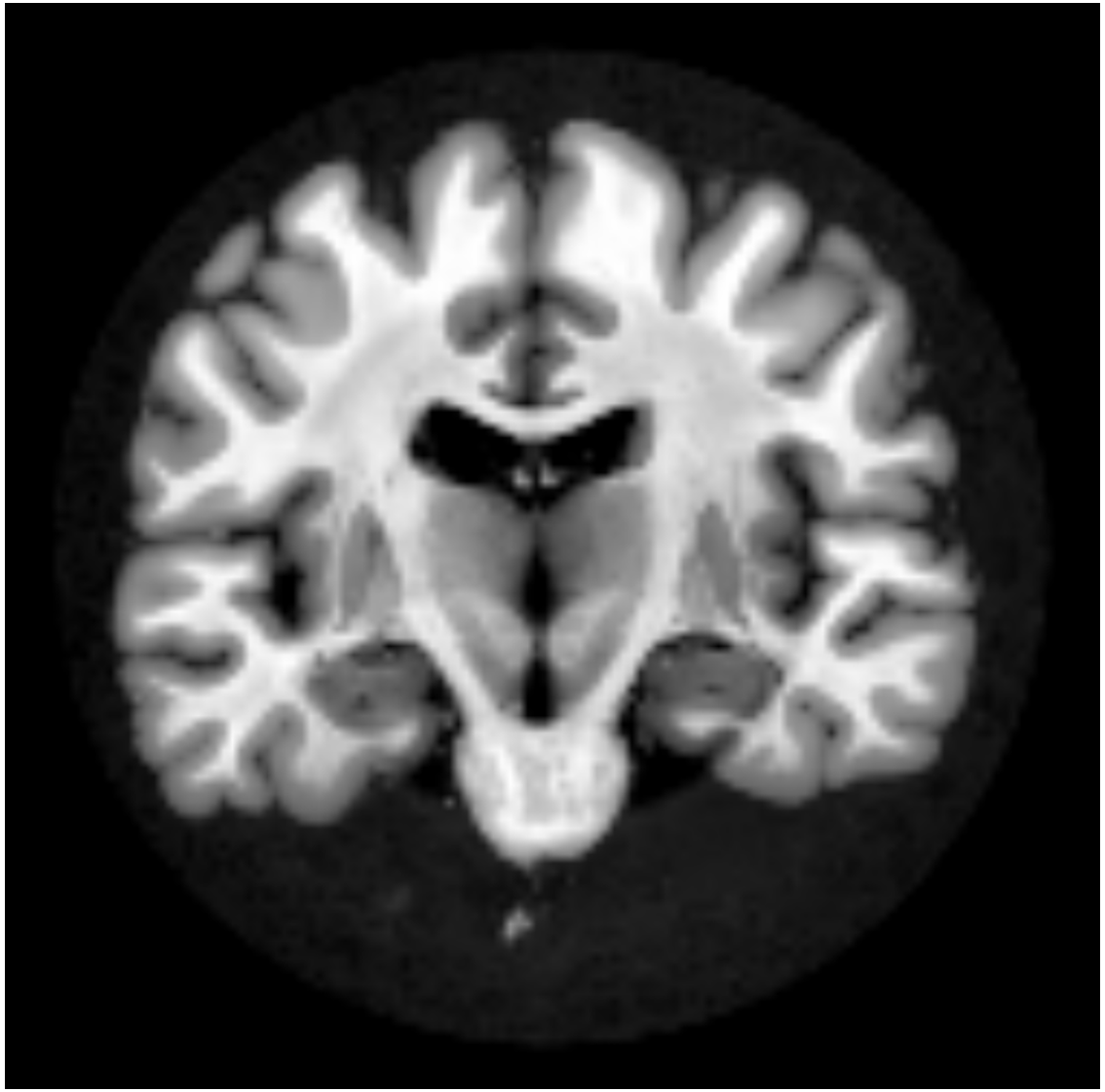}
    \caption{Brain 01}
\end{subfigure}
\hfill
\begin{subfigure}[b]{0.23\textwidth}
    \centering
    \includegraphics[width=\textwidth]{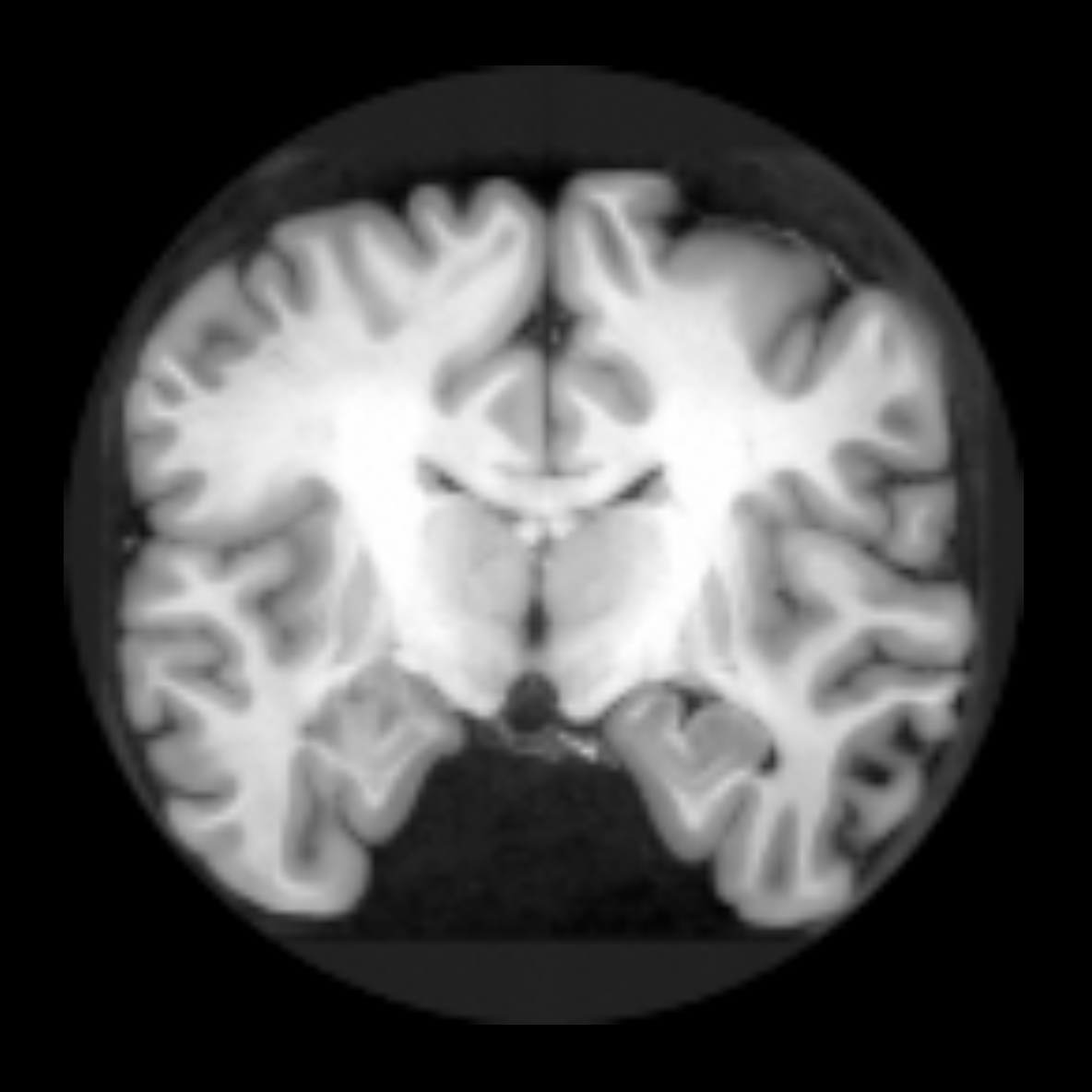}
    \caption{Brain 02}
\end{subfigure}
\hfill
\begin{subfigure}[b]{0.23\textwidth}
    \centering
    \includegraphics[width=\textwidth]{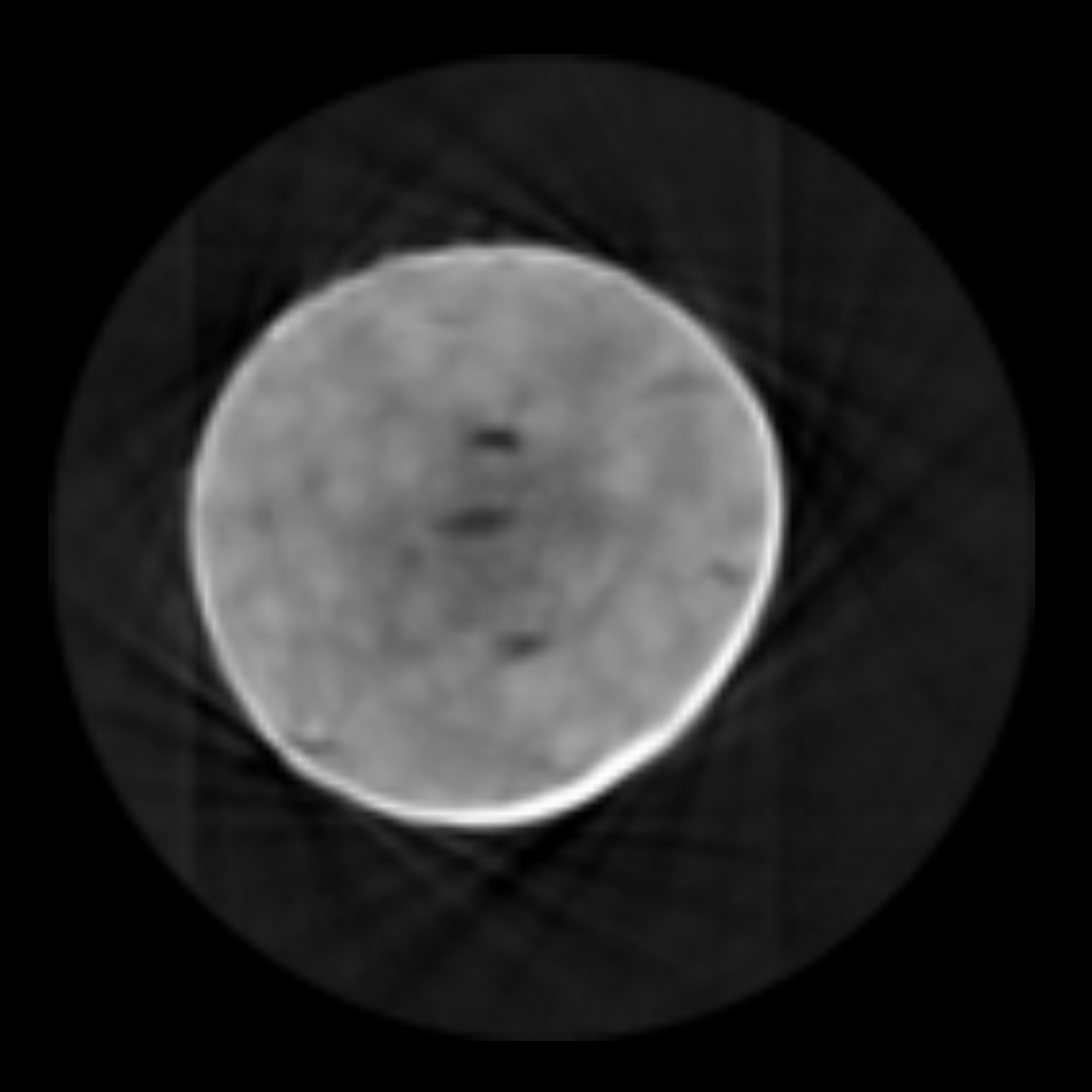}
    \caption{Elephant tusk 01}
\end{subfigure}
\hfill
\begin{subfigure}[b]{0.23\textwidth}
    \centering
    \includegraphics[width=\textwidth]{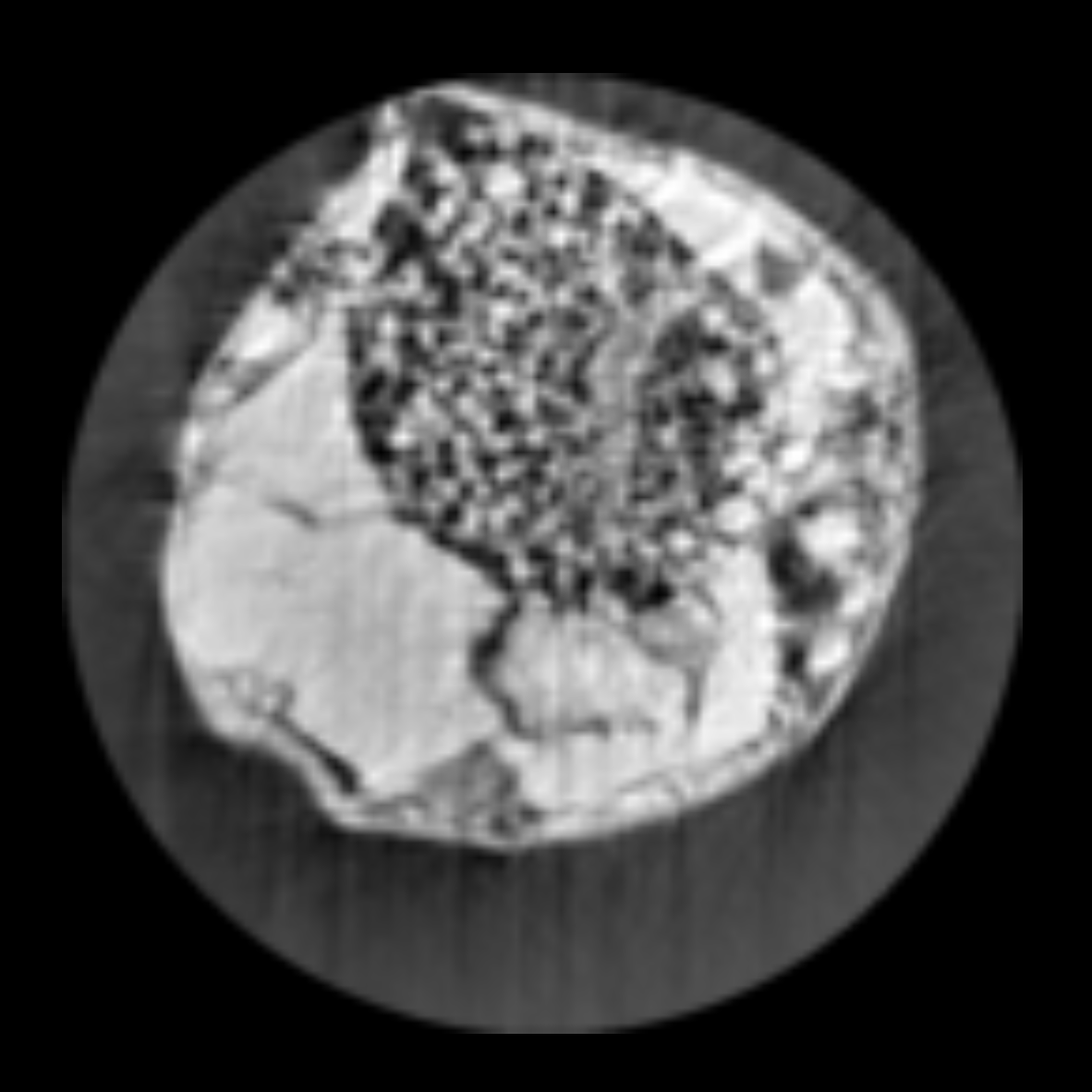}
    \caption{Worm 01}
\end{subfigure}
\caption{Phase-contrast \textmu CT volumes for eight representative samples, serving as ground truth.}
\label{fig:data_gallery}
\end{figure}

To ensure the universality of DL-VIF to arbitrary samples and morphological structures, we curated data from seventeen volumetric specimens: an elephant tusk, two worm samples, five human hearts, three human brains, four human kidneys, and one human testis, sourced from DESY (elephant tusk, worms) and the public ESRF repositories (heart, brain, kidney, testis) \cite{walsh2026humanorganatlas, esrf2026brain, esrf2025brain, walsh2024heart, walsh2024heart19, brunet2024heart, rahmani2024kidney, esrf2025kidney, HOA_Testis_2025, HOA_Heart_2026, Walsh2024Brain}. The input volumes used in this study vary in native resolution and voxel dimension from $240$ nm to $200$ \textmu m and $500^3$ to $1200^3$ voxels, reflecting differences in specimen size and acquisition protocol across the dataset. Figure \ref{fig:data_gallery} shows some example samples from the curated data gallery. Motion-corrupted counterparts of the curated volumes were synthesized by injecting jitter into the projection geometry during forward projection and reconstructing back, following the same jitter-matrix formulation used in Section \ref{secsimulation}.

Additional preprocessing includes a wide virtual detector ($1.6\times$ the volume diameter) to avoid truncation under large jitter, trilinear resampling of the reconstructed volume to $128^3$ voxels, and quantile-based intensity normalization (clipping at the 0.1/99.9 percentiles, then linear scaling to $[0,1]$). Jitter was simulated by using, for each of 4000 projection angles, an independent $4\times4$ rigid transformation combining a random translation ($\delta$ pixels, sampled uniformly in $[-\delta,\delta]$ per axis) and rotation ($\alpha$ degrees, sampled uniformly in $[-\alpha,\alpha]$). Cross-combining 20 translation levels ($\delta \in \{0,\dots,19\}$ px) with 7 rotation levels ($\alpha \in \{0^\circ,\dots,6^\circ\}$) produced 140 motion configurations per specimen. 

Applying jitter to all projection frames, however, produced a VIF distribution heavily skewed toward low quality values, since even a single-pixel jitter applied to every frame already reduces VIF drastically, leaving the high-VIF range (roughly $0.80$ to $1.0$) underrepresented. To fill this gap, a second, ratio-based jitter strategy was introduced in which motion is applied to only a randomly selected subset of frames (motion ratio $r \in \{0.1, 0.25, 0.5, 0.75, 0.9\}$), while the remaining frames stay static; combined with 10 translation levels and 3 rotation levels, this added 150 additional, more finely graded samples per specimen. 

Four (Worm~01, Heart~05, Brain~03, and Testis~01) out of the seventeen specimens are held out for testing. Note that the testis organ was not present throughout the training process, the Testis~01 is reserved for an out-of-distribution (OOD) test. Together, the two phases yield 290 motion-corrupted volumes per specimen and 3770 volumes in total for the thirteen training specimens, with VIF scores well distributed across the full $[0.2, 1.0]$ range. The mean and median of the VIF distribution on the training data pairs are $0.55$ and $0.53$, respectively. The generated training data pairs are split as follows: 2,900 pairs (290 per specimen) from ten specimens for training, 870 pairs from three specimens for validation. Each motion-corrupted volume's VIF label, computed against its motion-free reference, is precomputed and stored alongside the volume data. The VIF in this work is computed using a direct 3D extension of Sheikh and Bovik's local natural-scene-statistics VIF model \cite{Sheikh2006}, using a 4-level 3D Gaussian scale pyramid and a fixed additive-noise variance $\sigma_n^2 = 2.0$:
\begin{equation}
    \mathrm{VIF}_{3D}(P,T) = \frac{\displaystyle\sum_{j=0}^{3}\sum_{\mathbf{x}} \log_{10}\!\left(1 + \frac{g_j(\mathbf{x})^2\,\sigma_{T,j}^2(\mathbf{x})}{\sigma_{v,j}^2(\mathbf{x}) + \sigma_n^2}\right)}{\displaystyle\sum_{j=0}^{3}\sum_{\mathbf{x}} \log_{10}\!\left(1 + \frac{\sigma_{T,j}^2(\mathbf{x})}{\sigma_n^2}\right)} ,
\label{eq:vif3d}
\end{equation}
where $T$ and $P$ denote the (jointly normalized) reference and predicted volumes, $j$ indexes the four levels of a 3D Gaussian scale pyramid, $\sigma_{T,j}^2(\mathbf{x})$ is the local reference variance, $g_j(\mathbf{x})$ and $\sigma_{v,j}^2(\mathbf{x})$ are the local linear-channel gain and residual noise variance relating $P$ to $T$ at each voxel $\mathbf{x}$, and $\sigma_n^2$ is the fixed additive visual-noise variance of the assumed HVS channel.

Training is posed as a supervised regression problem: given a motion-corrupted volume, the network predicts a scalar DL-VIF score $\hat{V}$ approximating the reference-based VIF label $V$.  The two are compared with a weighted mean squared error loss over a batch of $N$ samples,
\begin{equation}
    L = \alpha \cdot \frac{1}{N}\sum_{i=1}^{N} (\hat{V}_i - V_i)^2,
\label{mse}
\end{equation}
with the scaling factor $\alpha = 2.0$ used to strengthen the gradient signal, particularly early in training.

The network was trained for 150 epochs on four NVIDIA RTX 4090 (24 GB) GPUs with a batch size of 16, using the Adam optimizer (initial learning rate $1\times10^{-6}$, weight decay $3\times10^{-5}$) and gradient clipping at a norm of 1.0. Automatic mixed precision was used throughout to reduce memory usage and accelerate training, and the learning rate was halved whenever the validation loss failed to improve for 10 consecutive epochs. 

\section{Results}\label{sec3}
This section reports the experimental outcomes of the various motion quality metrics and their application to geometry-based motion compensation. Section~\ref{sec:metrics_jitter} first evaluates how the candidate VIF metric responds to synthetically introduced jitter of increasing severity in comparison to other conventional metrics and thereby verifies the choice of VIF as the learning target. Section~\ref{sec:dlvif_training} examines the training behavior and predictive accuracy of the DL-VIF network, assessing how closely its reference-free score approximates the ground-truth VIF value on unseen specimens. Finally, Section~\ref{sec:optimization_result} applies the trained network within the gradient-based geometry optimization framework and reports the resulting jitter correction performance, comparing with the optimization results using other metrics.

\subsection{Metrics evaluation with controlled jitter levels}\label{sec:metrics_jitter}
The motion metric evaluation was carried out with the test specimen Worm~01. Table~\ref{tab:metrics_comparison} reveals a distinctive non-linear trend for the 3D VIF score as the six motion levels progress: an initially steep drop is followed by a flattening curve in which further increases in jitter magnitude yield progressively smaller reductions. This pattern can be traced back to how VIF is formulated around shared statistical information between reference and distorted volumes. Since high-frequency content in the reconstruction is the first to be disturbed by any positional offset, even a mild jitter level already removes a large share of it, producing the pronounced drop observed between the motion-free case and Level 1. Once most of this high-frequency information has been depleted, additional blurring introduced by stronger jitter has comparatively little left to erode, which explains why the score's decline slows markedly at the higher motion levels.

\begin{table}[ht!]
\centering
\begin{tabular}{|l|c|c|c|c|c|c|}
\hline
\textbf{Scenario} & \textbf{Trans [px]} & \textbf{Rot [$^\circ$]} & \textbf{Sharpness} & \textbf{TV} & \textbf{3D SSIM} & \textbf{3D VIF} \\ \hline
(a) No Motion   & 0       & 0     & 1.617 & 4.890 & 1.000 & 1.000 \\ \hline
(b) Level 1     & $\pm 2$ & 0     & 1.457 & 4.788 & 0.966 & 0.644 \\ \hline
(c) Level 2     & $\pm 5$ & $\pm 1$ & 1.252 & 4.660 & 0.757 & 0.309 \\ \hline
(d) Level 3     & $\pm 8$ & $\pm 2$ & 1.097 & 4.454 & 0.574 & 0.175 \\ \hline
(e) Level 4     & $\pm 10$ & $\pm 3$ & 1.025 & 4.250 & 0.510 & 0.134 \\ \hline
(f) Level 5     & $\pm 15$ & $\pm 4$ & 0.965 & 4.128 & 0.443 & 0.095 \\ \hline
\end{tabular}
\caption{Quantitative comparison of different image metrics across six simulated motion levels.}
\label{tab:metrics_comparison}
\end{table}

The 3D SSIM values in the same table follow a noticeably different trajectory, decreasing at a more constant rate across all six scenarios. Since SSIM is built around global structural correlation rather than a frequency-domain information measure, it responds more proportionally to the cumulative geometric displacement of the volume rather than being dominated by the earliest, most abrupt loss of fine detail. Taken together, the contrasting shapes of the two curves are consistent with what each metric is designed to capture: overall structural agreement for SSIM versus preserved information content for VIF.

The regression target VIF shows saturation in its response, while 3D SSIM and other metrics decline more uniformly, as shown in Table~\ref{tab:metrics_comparison}. As an optimization objective, it's beneficial to have high sensitivity near the optimum, and it is there that VIF is steepest; saturation occurs only in the regime where high-frequency content is already destroyed. More importantly, because VIF derives from a natural-scene-statistics model rather than local structural comparison~\cite{Sheikh2006}, it measures information loss largely independently of what the specimen looks like. For a surrogate that must transfer to unseen morphologies, this outweighs a linear response curve.

\subsection{DL-VIF training result}\label{sec:dlvif_training}
\begin{figure}[H]
    \centering
    \includegraphics[width=0.8\linewidth]{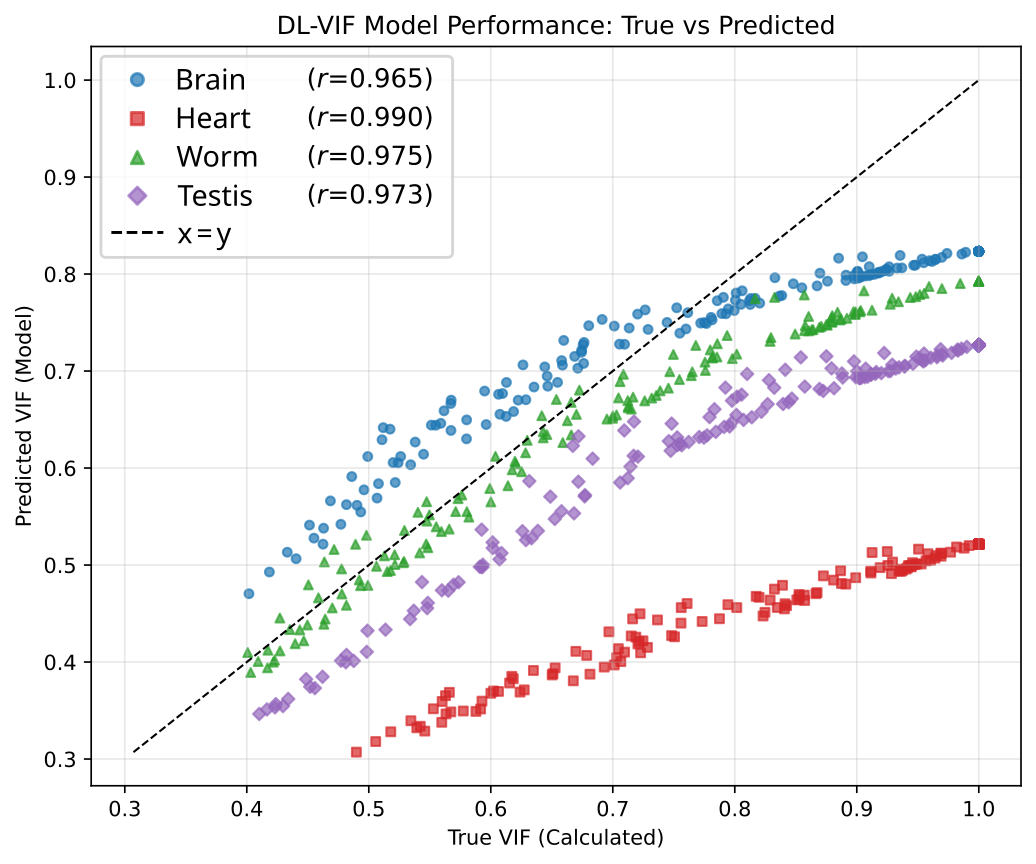}
    \caption{Predicted DL-VIF scores versus the analytically computed ground-truth VIF for four held-out specimens (Brain~05, Heart~05, Worm~01, and Testis~01) across a range of simulated motion severities (varying translation magnitude, rotation magnitude, and motion-duration ratio). Each marker corresponds to one simulated motion instance. The Pearson linear correlation coefficient ($r$) for each specimen is reported in the legend.}
    \label{fig:vif_scatter}
\end{figure}

The performance of the proposed CNN architecture was evaluated on four test specimens (Brain~05, Heart~05, Worm~01, and Testis~01). Figure \ref{fig:vif_scatter} illustrates the regression analysis between the predicted DL-VIF scores and the ground-truth VIF labels after 150 training epochs. 

The figure shows that, for every specimen, the predicted DL-VIF score increases monotonically and near-linearly with the analytically computed ground-truth VIF, yielding uniformly high Pearson linear correlation coefficients (PLCC) ($r = 0.965$ for brain specimen, $r = 0.990$ for heart specimen, $r = 0.975$ for worm specimen, and $r = 0.973$ for testis specimen). Since $r$ is invariant to affine rescaling of the data, these values confirm that the network reliably tracks the relative degradation in reconstruction quality induced by increasing jitter severity across markedly different anatomies. Notably, the heart specimen attains the highest correlation of the four specimens ($r = 0.990$), even though its predictions are visibly and systematically offset below the identity line across the entire true-VIF range. This combination of a near-perfect linear trend together with a large, roughly constant downward bias shows a strong precision (association, captured by $r$) but low accuracy (agreement with the identity line) \cite{Bland1986, Lin1989}. We attribute this offset to a distribution shift between the appearance statistics of the test specimens and those of the network's training set. Each volume is normalized independently via a global joint min-max rescaling prior to inference, so if a test specimen exhibits a different proportion of high-contrast structures (e.g., chamber walls, vasculature) or a different noise/texture profile than the training specimens, the effective intensity range and local variance statistics seen by the convolutional layers shift accordingly \cite{Geirhos2019}, biasing the scalar output of the regression head. Crucially, this offset need not be fatal for the downstream optimization: since gradient-based optimization is driven by the local slope of the loss surface along the search trajectory rather than its absolute value, a strong monotonic relationship between predicted and true VIF can still possibly steer the optimizer toward the correct optimization parameters even under a constant bias. Lastly, the strong linear correlation obtained for the completely OOD testis specimen ($r = 0.973$), an organ type entirely absent from the training set, is particularly encouraging, as it demonstrates the proposed model's potential to generalize its motion sensitivity to previously unseen anatomical structures.

\subsection{Optimization result}\label{sec:optimization_result}

To compare the performance of different image quality metrics for jitter correction and to evaluate the generalization capability of the proposed optimization algorithm across different samples, we conducted optimization experiments on the four held-out test specimens. The ground-truth projection data were simulated from volumes of the heart, brain, and testis specimens, while the projection data of the worm were acquired at the Hereon at DESY P05 nano-CT beamline. The target projection data was corrupted with simulated rigid jitter, with translation shifts sampled within $[-8, 8]$ pixels and rotations within $[-2^\circ, 2^\circ]$. For each metric and objective function (sharpness, sharpness with TV regularization, reference-based 3D SSIM, reference-based 3D VIF, DL-VIF, and DL-VIF with TV regularization), we evaluated the quality of the optimized reconstruction by calculating its SSIM relative to the motion-free ground-truth volume. 

All six modes share the same optimization scheme. Jitter is parameterized independently per projection angle as a 3-component translation and a 3-component Euler-angle rotation vector, giving $6N$ free scalar parameters for $N$ angles ($N=4000$), jointly updated by Adam optimizer with separate per-mode learning rates (Table~\ref{tab:opt_params}). The rotation learning rate is set roughly two orders of magnitude below the translation learning rate in every mode. Iteration count and learning rate were tuned per objective. Two modes, DL-VIF with TV regularization and sharpness with TV regularization, additionally penalize a background total-variation term (masked at the 65th intensity percentile) to suppress artifacts outside the imaged object, weighted against the primary quality term as shown in the table.

\begin{table}[t]
\centering
\begin{tabular}{llccll}
\toprule
Objective & Iter. & LR (transl.) & LR (rot.) & Loss weights / notes \\
\midrule
Sharpness     & 100 & 0.030 & 0.00030 & Blind sharpness \\
Sharpness $+$ TV reg.         & 100 & 0.030 & 0.00030 & $\beta=3$; mask at 65th percentile \\
SSIM        & 150 & 0.050 & 0.00050 & SSIM vs.\ GT \\
VIF      & 150 & 0.100 & 0.00100 & Analytic VIF vs.\ GT \\
DL-VIF            & 160 & 0.060 & 0.00060 & DL-VIF only, no regularizer \\
DL-VIF $+$ TV reg.         & 160 & 0.060 & 0.00060 & $\beta=0.25$; mask at 65th percentile \\
\bottomrule
\end{tabular}
\caption{Default optimization hyperparameters for each objective-function mode.
LR values are Adam learning rates.}
\label{tab:opt_params}
\end{table}

\begin{figure}[H]
\centering
\includegraphics[width=1.0\textwidth]{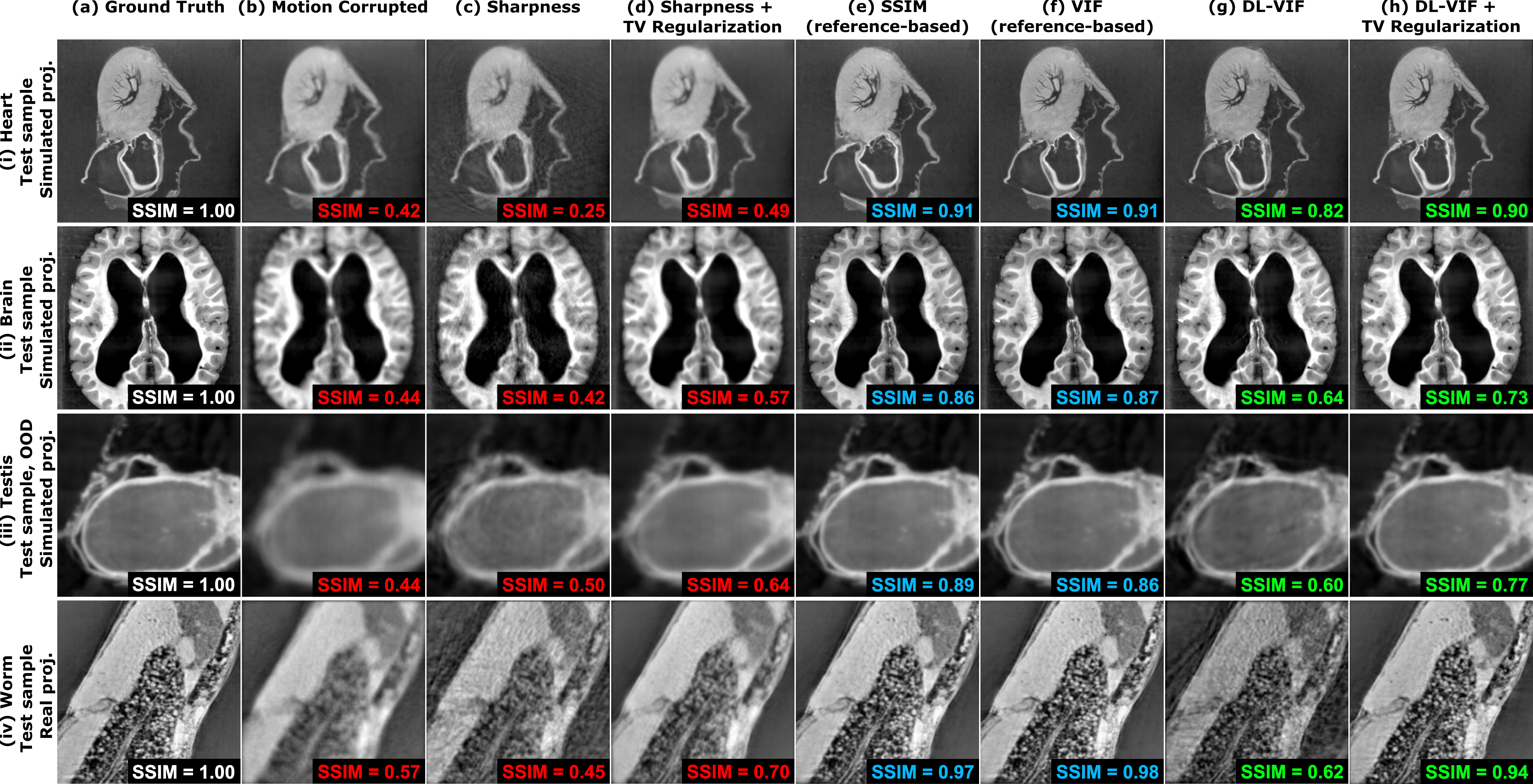}
\caption{Comparison of motion-correction results across four representative test cases. (i) and (ii) are in-distribution organs (heart, brain) using simulated target projection with jitter motion. (iii) is an OOD organ (testis) using simulated target projection with jitter motion. (iv) is a real projection acquisition of a worm specimen with manually introduced jitter motion. Columns show (a) the artifact-free ground truth, (b) the uncorrected motion-corrupted input, and optimized reconstructions obtained by optimizing six different objectives: (c) sharpness maximization, (d) sharpness with TV regularization, (e) SSIM and (f) analytic VIF computed directly against the ground truth (reference-based upper-bound baselines), (g) the proposed DL-VIF network, and (h) the proposed DL-VIF with TV regularization. The SSIM score relative to the ground truth is displayed in red on each image.}
\label{fig:full_comparison}
\end{figure}

Figure~\ref{fig:full_comparison} compares reconstructions obtained by optimizing six different objective functions against the ground-truth and motion-corrupted baselines for four test specimens. Reference-based SSIM and VIF optimization (columns e--f), which have direct access to the ground-truth volume during optimization, consistently achieve the highest SSIM scores across all four cases (e.g., 0.9108/0.9079 for the heart and 0.9731/0.9799 for the worm), serving as a practical upper bound on the reconstruction quality achievable by this pipeline. In contrast, blind sharpness maximization (column c) proves unreliable on its own: for three of the four specimens (heart, brain, worm) it actually reduces SSIM below the uncorrected motion-corrupted baseline (e.g., 0.2476 vs.\ 0.4247 for the heart), indicating that sharpening alone can converge to sharp but geometrically incorrect solutions, such as amplified noise or streak artifacts, rather than the true motion-free structure. Adding TV regularization to the sharpness objective (column d) substantially mitigates this failure mode, yielding a consistent improvement over the motion-corrupted baseline for every specimen. The proposed DL-VIF network (column g), which, unlike SSIM/VIF, requires no ground-truth access at inference time, likewise produces consistent gains over the uncorrected baseline, and combining it with TV regularization (column h) further closes the gap toward the reference-based upper bound in every case (e.g., 0.6008 to 0.7664 for the OOD testis specimen, and 0.6183 to 0.9411 for the worm specimen with real projection). Note that the heart specimen, for which DL-VIF+TV correction (SSIM $= 0.8972$) nearly matches the reference-based upper bound (SSIM $=0.9079$), even though the same specimen exhibited one of the largest systematic offsets between predicted and true VIF observed in Figure~\ref{fig:vif_scatter} ($r = 0.990$). This empirically confirms the argument made in the previous section that it is the strong monotonic correlation between predicted and true VIF that allows gradient-based optimization to be reliably navigated toward the correct motion parameters despite the offset. Notably, the testis organ (row iii) was never seen during DL-VIF network training, yet DL-VIF+TV correction (0.7664) approaches the quality achieved by the ground-truth-informed SSIM and VIF optimizations (0.8892 and 0.8626, respectively), supporting the model's ability to generalize its learned motion-sensitivity to an unseen organ. Finally, the worm case (row iv) demonstrates that these gains transfer from simulated projections to real, motion-corrupted acquisitions: the DL-VIF+TV objective achieves an SSIM of 0.9411, comparable to the reference-based methods (0.9731 and 0.9799).

\section{Discussion and conclusion}\label{sec4}

We have made auto-focus geometry correction practical for jitter correction of parallel-beam phase-contrast \textmu CT through three components: an analytic differentiation of the reconstruction operator under parallel projection, a compact 3D CNN that predicts the 3D VIF score of a single volume (PLCC $>0.96$ on held-out specimens), removing the need for a motion-free reference at deployment, and a normalized TV penalty restricted to the image background that keeps the optimization physically constrained. Across all four test cases that span two in-distribution organs (heart, brain), one out-of-distribution specimen never seen during network training (testis), and one real, non-simulated acquisition (worm), the pipeline recovers structural detail lost to per-projection jitter without any pre-scan reference, raising the SSIM against the motion-free ground truth from 0.42--0.57 uncorrected to 0.73--0.94 after correction (e.g., 0.9411 vs.\ 0.5740 for the real worm acquisition). These results indicate that a differentiable geometry gradient coupled to a deliberately structure-agnostic learned quality metric, with a spatially selective regularizer to keep the optimization physically constrained, is a viable route to reference-free jitter correction in phase-contrast \textmu CT across varying morphologies and imaging conditions.

Transferability is where our setting departs most from prior learned motion metrics. Huang et al.~\cite{Huang2022} and Thies et al.~\cite{Thies2025} both address human head CT, where fixed anatomy permits a network to exploit strong shape priors. No such prior exists at a \textmu CT beamline scanning arbitrary research samples. We addressed this through the data, curating specimens across five morphological classes, and through the architecture, compressing the feature map to $2\times2\times2$ before regression. The architecture modification departs from Huang et al., who retain spatial resolution to localize regional artifacts; since jitter degrades the volume globally, our network needs only to report an overall degradation level, and discarding positional information shrinks the first fully connected layer by roughly four orders of magnitude.

An unanticipated finding is how readily a gradient-based optimizer exploits a learned quality term, discovering geometries that raise the predicted DL-VIF score while degrading the reconstruction by populating empty regions with texture the network has never seen in training. This is a consequence of optimizing against a surrogate outside its fitted distribution and should be expected of such a gradient-driven learned auto-focus criterion. Restricting the TV penalty to the background is an effective remedy since it constrains only where it is known to be featureless, avoiding the loss of genuine internal structure that a global TV term causes.

Several limitations are yet to be further studied. First, achievable image quality is bounded by the ground-truth volumes used to generate the training labels, and the preprocessing resamples each volume to $128^3$ voxels, which efficiently reduces the memory footprint during inference and optimization, but also discards precisely the high-frequency content that the metric should detect. The optimizer cannot recover details of its objective that it is blind to, so this plausibly caps correction accuracy. Quantifying the ceiling by training at progressively higher input resolutions is worth checking in the future study. Second, the evaluation relies on jitter injected into forward-projected data, so the assumed motion model matches the one used to generate it. Validation against measured jitter with independent tracking~\cite{Brombal2021-ho} would close this gap, as would benchmarking against the projection-alignment methods established in the synchrotron community. Finally, other geometry error sources, such as steady sample drift and incorrect setup calibration, might result in artifacts that are visually different from the jitter artifact on which the DL-VIF network was trained, and its generalizability to these unseen error types remains to be verified. Future work should therefore investigate whether retraining on an augmented dataset covering a broader range of geometric error sources, further architectural adaptation of the network, or a purely mathematically derived, CT-geometry-based motion metric would be needed to extend the proposed framework beyond jitter correction.

\section{Acknowlegdement}\label{sec5}
We acknowledge Helmholtz-Zentrum Hereon (Geesthacht, Germany) and DESY (Hamburg, Germany) for providing the experimental facilities. Part of the dataset in this research was collected at PETRA~III. Beamtime was allocated for proposal I-20221419 and I-20240219. We thank Sara Savatović for recommending the public dataset for this study. Furthermore, we thank Martin Dierolf for the support on the IT issues during the experiment.


\end{document}